\documentclass{article}
\usepackage{ijcai18}

\usepackage{graphicx}
\usepackage{epstopdf}
\usepackage{times}
\usepackage{xcolor}
\usepackage{soul}
\usepackage[utf8]{inputenc}
\usepackage[small]{caption}
\usepackage{times}
\usepackage{amsmath}
\DeclareMathOperator*{\argmax}{argmax}
\usepackage{latexsym}
\usepackage{balance}
\usepackage{url}
\usepackage{amsfonts}
\usepackage{algorithm} 
\usepackage{algorithmic}  
\usepackage[algo2e,linesnumbered]{algorithm2e} 
\usepackage{amsfonts}
\usepackage{amsbsy}
\usepackage{physics}
\usepackage{color}
\usepackage{subfigure}
\usepackage{booktabs}
\usepackage{multirow}
\usepackage{paralist}
\usepackage{dsfont}
\usepackage{booktabs}

\newcommand{\Loss}{\mathcal{L}}

\title{MEGAN: Mixture of Experts of Generative Adversarial Networks\\ 
for Multimodal Image Generation}

\author{
David Keetae Park$^\dagger$, 
Seungjoo Yoo$^\dagger$, 
Hyojin Bahng$^\dagger$, 
Jaegul Choo$^\dagger$,
Noseong Park$^\ddagger$
\\ 
$^\dagger$ Korea University, 
\{heykeetae, seungjooyoo, hjj552, jchoo\}@korea.ac.kr\\
$^\ddagger$ University of North Carolina at Charlotte,
npark2@uncc.edu
}
\begin{document}
\maketitle

\begin{abstract}
\let\thefootnote\relax\footnotetext{$\ddagger$ Corresponding author; This work was supported by the National Research Council of Science \& Technology (NST) grant by the Korea government (MSIP) [No. CRC-15-05-ETRI].}Recently, generative adversarial networks (GANs) have shown promising performance in generating realistic images. However, they often struggle in learning complex underlying modalities in a given dataset, resulting in poor-quality generated images. To mitigate this problem, we present a novel approach called \textit{mixture of experts GAN} (MEGAN), an ensemble approach of multiple generator networks. Each generator network in MEGAN specializes in generating images with a particular subset of modalities, e.g., an image class. Instead of incorporating a separate step of handcrafted clustering of multiple modalities, our proposed model is trained through an end-to-end learning of multiple generators via gating networks, which is responsible for choosing the appropriate generator network for a given condition. We adopt the categorical reparameterization trick for a categorical decision to be made in selecting a generator while maintaining the flow of the gradients. We demonstrate that individual generators learn different and salient subparts of the data and achieve a multiscale structural similarity (MS-SSIM) score of 0.2470 for CelebA and a competitive unsupervised inception score of 8.33 in CIFAR-10.
\end{abstract}

\section{Introduction}
Since the introduction of generative adversarial networks (GANs) \cite{goodfellow2014generative}, researchers have dove deeply into improving the quality of generated images. Recently, a number of new approaches have been proposed for high-quality image generation, e.g., ProgressiveGAN~\cite{karras2017progressive}, SplittingGAN~\cite{grinblat2017class}, SGAN~\cite{huang2017stacked}, and WGAN-GP~\cite{salimans2016improved}. 

\begin{figure}[t]
\centering
\includegraphics[width=0.5\textwidth]{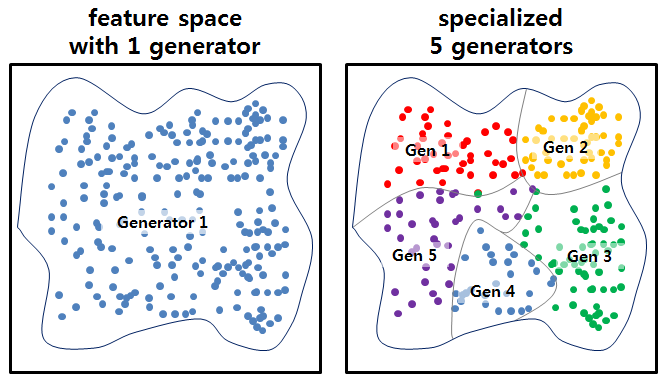}
\caption{Multiple generators specialized in particular data clusters}\label{fig:intro}
\end{figure}

We propose a novel GAN model equipped with multiple generators, each of which specializes in learning a certain modality of dataset (see Fig.~\ref{fig:intro}). In addition to the generators, we employ an auxiliary network that determines a generator that will be trained from a certain training instance. We name the auxiliary network as \textit{gating networks} following the precedent~\cite{jacobs1991adaptive}. 

Ensembling multiple neural networks coupled with gating networks was first introduced to achieve a higher performance in multi-speaker phoneme recognition~\cite{hampshire1990meta}. In their method, the design of the loss function caused the neural networks to cooperate. A later research introduced a new loss function that stimulates competitions among neural networks, where the involved neural networks attempt to specialize in a certain task rather than redundantly learn the same feature~\cite{jacobs1991adaptive}. The algorithm is now called \textit{mixture of experts} in various machine learning domains. Reminiscent of their work, we name our proposed GAN approach as MEGAN, short for the mixture of experts GAN.
 
 The gating networks in our proposed MEGAN are responsible for selecting one particular generator that would perform best given a certain condition. The gating networks consist of two submodules, an \textit{assignment module} and \textit{Straight-Through Gumbel-Softmax}~\cite{jang2016categorical}, which we will discuss in detail in Section~\ref{sec:gating}.
 
 Although MEGAN inherits the idea of multiple generators and the gating networks, we will not adopt the proposed loss function~\cite{jacobs1991adaptive} but utilize adversarial learning to leverage the latest success of GANs.
 
 Our work has two contributions. First, we build a mixture of experts GAN algorithms that are capable of encouraging generators to learn different modalities existing in our data. Second, we utilize the newly discovered Gumbel-Softmax reparameterization trick and develop the regularization for \textit{load-balancing} to further stabilize the training of MEGAN. We evaluate our model using various criteria, notably achieving an MS-SSIM score of 0.2470 for CelebA, which suggests that MEGAN generates more diverse images compared to other baseline models. Our generated samples also achieve a competitive inception score of 8.33 in an unsupervised setting.

\section{Related Work}

Several studies on GANs have been proposed to stabilize the learning process and improve the quality of generated samples. Some of these studies incorporated novel distance metrics to achieve better results. For instance, the original GAN~\cite{goodfellow2014generative} suffers from the vanishing gradient problem arising from the sigmoid cross-entropy loss function used by the discriminator. LSGAN~\cite{mao2017least} solves this problem by substituting the cross-entropy loss with the least-squares loss function. WGAN~\cite{arjovsky2017wasserstein} adopts the Earth mover's distance that enables an optimal training and solves the infamous mode collapse. WGAN-GP progresses one step further by adopting a gradient penalty term for a stable training and higher performance. Meanwhile, BEGAN~\cite{berthelot2017began} aims to match auto-encoder loss distributions using a loss elicited from the Wasserstein distance, instead of matching the data distributions directly. In addition, DRAGAN~\cite{kodali2017train} prevents mode collapse using a no-regret algorithm.

 Other algorithms such as AdaGAN~\cite{tolstikhin2017adagan} and MGAN~\cite{hoang2017multi} employ ensembling approaches, owing to having multiple generators to learn complex distributions. Based on the idea of boosting in the context of the ensemble model, AdaGAN trains generators sequentially while adding a new individual generator into a mixture of generators. While AdaGan gradually decreases the importance of generators as more generators are added into the model, within the framework of our proposed MEGAN, we pursue an equal balancing between generators by explicitly regularizing the model, which avoids the problem of being dominated by a particular generator.
 
MGAN adopts a predefined mixture weight of generators and trains all generators simultaneously; however, our proposed MEGAN dynamically alters generators through gating networks and train the generators one at a time. The MGAN's fixed mixture model is sub-optimal, compared to our trainable mixture model.
 Our proposed MEGAN is different from these models in that each generator can generate images on its own and learn different and salient features.
 
MAD-GAN~\cite{ghosh2017multi} has strong relevance to our work. Having multiple generators and a carefully designed discriminator, MAD-GAN overcomes the mode-collapsing problem by explicitly forcing each generator to learn different mode clusters of a dataset. Our MEGAN and MAD-GAN are similar in that both models allow the generators to specialize in different submodalities. However, MEGAN is differentiated from MAD-GAN in two aspects. First, all generators in MEGAN share the same latent vector space, while the generators of MAD-GAN are built on separated latent vector spaces. Second, the generators of MAD-GAN can theoretically learn the identical mode clusters; however, the gating networks built in our MEGAN ensure that each generator learns different modes by its design.

\section{Categorical Reparameterization}
\label{sec:categorical}

Essentially, GANs generate images when given latent vectors. Given $n$ generators and a latent vector $\mathbf{z}$, our model aims to select a particular generator that will produce the best-quality image. It essentially raises the question as to how a categorical decision is made. The Gumbel-Max trick~\cite{gumbel1954statistical,maddison2014sampling} allows to sample a one-hot vector $\mathbf{s}$ based on the underlying probability distribution $\pi_i$:
\begin{align}
\begin{split}
\mathbf{s} &= one\_hot(\argmax_{i}[a_{i}+\log\pi_{i}]) \\ 
a_{i} &= -\log({-\log(u_{i})})
\end{split}\end{align}
where $u_i$ is sampled from Uniform(0,1). However, the $\argmax$ operator in the Gumbel-Max trick is a stumbling block when training via back propagation because it gives zero gradients passing through the stochastic variable and precludes gradients from flowing further. A recent finding~\cite{jang2016categorical} suggests an efficient detour to back propagate even in the presence of discrete random variables by a categorical reparameterization trick. The Gumbel-Softmax function generates a sample $\mathbf{y}$ that approximates $\mathbf{s}$ as follows:
\begin{equation}
\begin{split}
\mathbf{y}_i &= \frac{\exp((\log\pi_i+a_i)/\tau)}{\sum_{j=1}^k\exp((\log\pi_j+a_j)/\mathbf{\tau})}
\end{split}\end{equation}
where $\mathbf{y}_i$ is an $i$-th component of the vector $\mathbf{y}$, and  $\mathbf{\tau}$ is the temperature that determines how closely the function approximates the sample $\mathbf{s}$. It is noteworthy that in practice, we directly predict $\log\pi_i$ through the \textit{assignment module} that we will discuss in Section~\ref{sec:gating}.     

\subsection{Straight-Through Gumbel-Softmax}
The Gumbel-Softmax method approximates the discrete sampling by gradually annealing the temperature $\tau$. It appears to be problematic in our setting because when the temperature is high; a Gumbel-Softmax distribution is not categorical, leading all generators to be engaged in producing a fake image for a given latent vector $\mathbf{z}$. Our objective is to choose the most appropriate generator. Therefore, we do not use the Gumbel-Softmax but adopt the following Straight-Through Gumbel-Softmax (STGS).

The STGS always generates discrete outputs (even when the temperature is high) while allowing the gradients flow. In practice, the STGS calculates $\mathbf{y}_i$ but returns $\mathbf{y\textsubscript{hard}}$:
\begin{equation}\label{eq:stgs}
\begin{split}
\mathbf{y\textsubscript{hard}} &= \mathbf{y}_i + (one\_hot(\argmax_{i}(\mathbf{y}_i)) - \mathbf{y}_d)
\end{split}\end{equation}
where $\mathbf{y}_d$ is a variable having the same value as $\mathbf{y}_i$ but is detached from the computation graph. With this trick, the gradients flow through $\mathbf{y}_i$ and allows the networks to be trained with the annealing temperature. 

\begin{figure*}[t]
\centering
\includegraphics[width=0.98\textwidth]{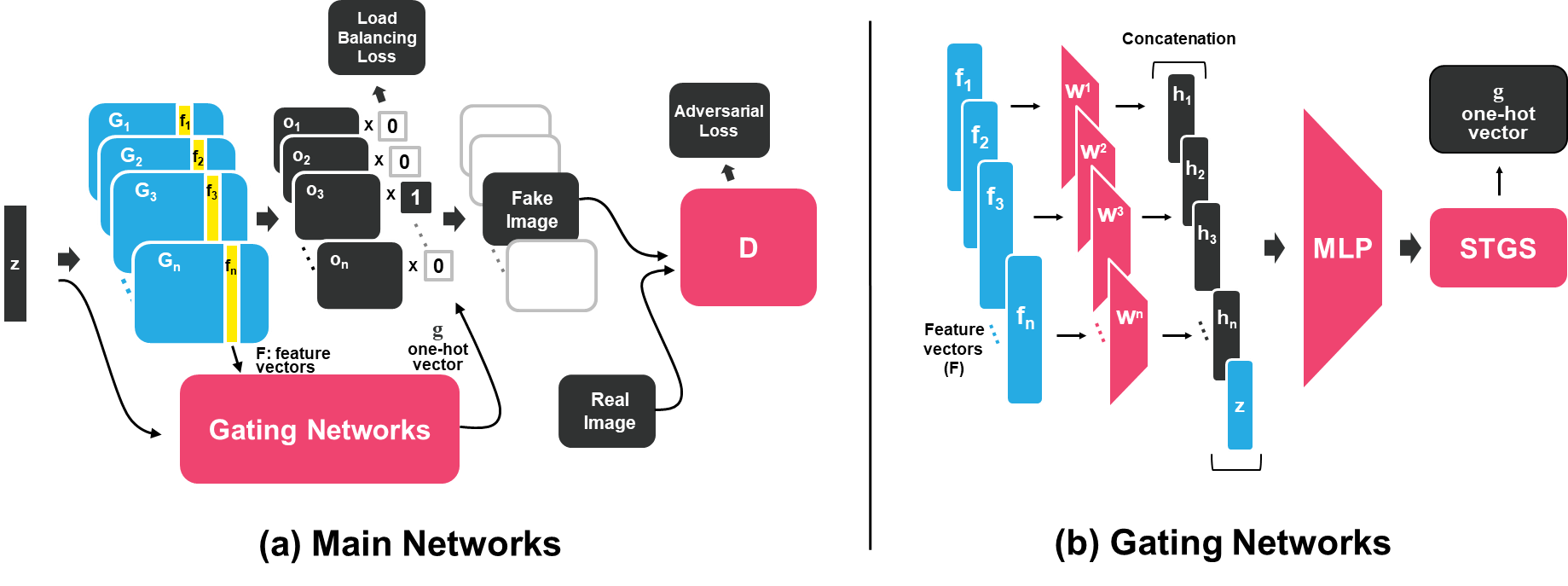}
\caption{\textbf{The proposed architecture of MEGAN}; \textbf{(a)} shows the overview of our main networks. Given a latent vector $\mathbf{z}$, each of the $n$ generators produces an output ${o_i}$. The latent vector \textbf{z} and $n$ feature vectors (denoted in yellow) extracted from the generators are given as input to the gating networks that produce a one-hot vector $\mathbf{g}$, as shown in the middle. The chosen image by the one-hot vector (marked as ``Fake Image") will be fed into the discriminator that measures the adversarial loss with regard to both real and fake classes. \textbf{(b)} illustrates an in-depth view on the gating networks. The gating networks output a one-hot vector $\mathbf{g}$.}\label{fig:mainnet}
\end{figure*}
 
\section{Mixture of Experts GAN}

In this section, we illustrate the details of our proposed MEGAN and discuss how generators become specialized in generating images with particular characteristics following the notion of the mixture of experts~\cite{jacobs1991adaptive}. 

\subsection{Proposed Network Architecture}
Let ${G_i}$ denote a generator in a set $\mathbf{G}=\{{G_1}, {G_2}, \cdots, {G_n}$\}, and $\mathbf{z}\sim \mathcal{N}(0,1)$ is a random latent vector. A latent vector $\mathbf{z}$ is fed into each generator, yielding images $\mathbf{o}_i\in \mathbf{O} = \{\mathbf{o}_1, \mathbf{o}_2, \cdots, \mathbf{o}_n$\} and their feature vectors $\mathbf{f}_i\in \mathbf{F} = \{\mathbf{f}_1, \mathbf{f}_2, \cdots, \mathbf{f}_n$\}. Each feature vector is produced in the middle of ${G_i}$. In our experiments, we particularly used the ReLU activation map from the second transposed convolution layer of the generator as the representative feature vector. The latent vector $\mathbf{z}$ and all of the feature vectors $\mathbf{f}_i$ are then passed to the gating networks to measure how well $\mathbf{z}$ fits each generator. The gating networks produce a one-hot vector $\mathbf{g} = \langle{g_1},{g_2}, \cdots, {g_n}\rangle$ where ${g_i} \in \{0,1\}$. We formulate the entire process as follows:
\begin{gather}
(\mathbf{f}_i, \mathbf{o}_i) = \mathit{G}_i(\mathbf{z})\label{eq:gen_feature}\\
\mathbf{g} = \textbf{GN}(\mathbf{z}, \mathbf{f}_1, \mathbf{f}_2, \cdots, \mathbf{f}_n)\label{eq:}\\
\mathbf{FI} =\sum_{i=1}^{n}{g}_i\mathbf{o}_i\label{eq:mainnet}
\end{gather}
where $\mathbf{FI}$ denotes the generated fake image that will be delivered to the discriminator, and \textbf{GN} is the gating network. Fig.~\ref{fig:mainnet} provides an overview of the proposed networks.

\subsection{Gating Networks}\label{sec:gating}
In the context of the mixture of experts, the gating networks play a central role in specialization of submodules~\cite{jacobs1991adaptive}. We use auxiliary gating networks that assign each $\mathbf{z}$ to a single generator so as to motivate each generator to learn different features. Concretely, we aim to train each generator ${G_i}$ to (1) be in charge of the images with certain characteristics potentially corresponding to a particular area in the entire feature space, and (2) learn the specialized area accordingly. The gating networks consist of two distinctive modules, an \textit{assignment module} that measures how well the latent vector $\mathbf{z}$ fits each generator and an \textit{STGS module} that samples a generator based on the underlying distribution given by the assignment module. 

\paragraph{Assignment Module}
The assignment module uses the feature vectors $\mathbf{f}_i\in\mathcal{R}^{k}$ and first encodes each of them into a hidden state $\mathbf{h}_i\in\mathcal{R}^{m}$ in a smaller dimension ${m}$:
\begin{equation}
\begin{split}
\mathbf{h}_i =ReLU(W^i\mathbf{f}_i)
\end{split}\end{equation}
where $W^i$ denotes a linear transformation for feature vector $\mathbf{f}_i$. Encoding each feature vector reduces the total complexity significantly, because $k$, the dimension of a feature map $\mathbf{f}_i$, is typically large, e.g., $k$ = 8192 in our implementation. The reduced dimension $m$ is a hyperparameter that we set as 100.  

The $\mathbf{h}_i$ are then concatenated along with the latent vector. The merged vector is then passed to a three-layer perceptron, which consists of batch normalizations and ReLU activations:
\begin{equation}
\begin{split}
\textit{\textbf{l}} = MLP([\mathbf{z}, \mathbf{h}_1, \mathbf{h}_2, \dots, \mathbf{h}_n])
\end{split}\end{equation}
where the resulting $\textit{\textbf{l}}\in\mathcal{R}^{n}$ is a logit vector, an input for the STGS. $\textit{\textbf{l}}$ also corresponds to $\log\pi_i$ explained in Section~\ref{sec:categorical}.  

\paragraph{STGS Module}
\textit{\textbf{l}} is an unnormalized density that determines the generator that most adequately fits the latent vector. The STGS samples a one-hot vector with \textit{\textbf{l}} as an underlying distribution. We denote the sampled one-hot vector as $\mathbf{g}$, which corresponds to y\textsubscript{hard} illustrated in Eq.~\eqref{eq:stgs}.
It strictly yields one-hot vectors. Thus, with the STGS, we can select a particular generator among many, enabling each generator to focus on a sub-area of the latent vector space decided by the gating networks. It is noteworthy that the assignment module is updated by the gradients flowing through the STGS module.


\subsection{Load-Balancing Regularization}
We observed that the gating networks converge too quickly, often resorting to only a few generators. The networks tend to be strongly affected by the first few data and favor generators chosen in its initial training stages over others. The fast convergence of the gating networks is undesirable because it leaves little room for other generators to learn in the later stages. Our goal is to assign the data space to all the generators involved. 

To prevent the networks from resorting to a few generators, we force the networks to choose the generators in equal frequencies in a mini-batch. Thus, we introduce a regularization to the model as follows:
\begin{align}\label{load_balance}
\begin{split}
\Loss_{LB}  &= \sum_{i=1}^n\norm{p_i-\frac{1}{n}}_2^2, \\
p_i &= \frac{\sum_{j=1}^b\mathds{1}[g_{i}^j=1]}{b},
\end{split}\end{align}
where $\Loss_{LB}$ indicates \textit{the load-balancing} loss, $b$ is the mini-batch size, ${g_{i}^j}$ is the $i$-th element of the one-hot vector $\mathbf{g}$ for the $j$-th data of a training mini-batch. $\mathit{p_i}$ is the probability that a certain generator will be chosen. The indicator function $\mathds{1}[g_{i}^j=1]$ returns 1 if and only if $g_{i}^j=1$. Concretely speaking, we train the model with mini-batch data; further, for all the data in a mini-batch, we count every assignment to each generator. Thus, the regularization loss pushes $\mathbf{g}$ to equally select generators.
  
\begin{algorithm2e}[t]
\DontPrintSemicolon
\caption{\strut Mini-batch training algorithm of MEGAN.\label{alg:gan}}
\hrule
\KwIn{Real Samples: $\{x_1, x_2, \cdots\}$; Mini-batch Size: $m$}
\KwOut{Generators: $\{G_1, \cdots, G_n\}$}
\hrule
$G_1, G_2, \cdots, G_n \gets$ $n$ generative neural networks\;
$D \gets$ one discriminator neural networks\;
$\lambda \gets$ a weight for \textit{the load-balancing} regularization \;
\While{until converge} {
\textit{iter} $\gets$ 0\;
$X \gets \{x_1, \cdots, x_m\}$, a mini-batch of real samples\;
$Z \gets \{\mathbf{z}_1, \cdots, \mathbf{z}_m\}$, a mini-batch of latent vectors\;
$\mathbf{\tau} \gets 0.5\exp^{-0.001\times{\textit{iter}}}$\;

\For{each $\mathbf{z}_i \in Z$} {

\For{j in $(1,2,\cdots,n)$}{
$\mathbf{f}_{i,j}$, $\mathbf{o}_{i,j}$ = $G_j(\mathbf{z}_i)$\;
}

$\textit{\textbf{l}}_i \gets AssignModule(\mathbf{z}_i, \mathbf{f}_{i,1}, \mathbf{f}_{i,2}, \cdots, \mathbf{f}_{i,n})$\; 
$\mathbf{g}_i = \langle\mathit{g_{i,1}}, \cdots, \mathit{g_{i,n}}\rangle =  STGS(\textit{\textbf{l}}_i, \mathbf{\tau})$\;  
Generate a fake image $\mathbf{FI}_i$ by $\sum_{j=1}^{n}\mathit{g}_{i,j}\mathbf{o_{i,j}}$\;
}

Train the discriminator $D$ using $\Loss_{adv}$\;
Train the generators $G_1, \cdots, G_n$ using $\Loss_{adv}$ \;
Train the gating networks using $\Loss_{adv}$ + $\lambda\Loss_{LB}$ \;
\textit{iter} += 1
}

\Return $\{G_1, G_2, \cdots, G_n\}$\;
\hrule
\end{algorithm2e}

\subsection{Total Loss}
The total loss of our model set for training is as follows:
\begin{equation}\label{load_balance}
\begin{split}
\Loss = \Loss_{adv} + \lambda\Loss_{LB}
\end{split}\end{equation}
where $\Loss_{adv}$ is any adversarial loss computed through an existing GAN model. We do not specify $\Loss_{adv}$ in this section, because it may vary based on the GAN framework used for training. We set $\lambda$ to control the impact of the load-balancing regularization.  

\subsection{Discussions}
In this chapter, we discuss a couple of potential issues about some difficulties in the mixture model.

\paragraph{Mechanism of Specialization} In MEGAN, what forces the generators to specialize? We presume it is the implicit dynamics between multiple generators and the STGS. No explicit loss function exists to teach the generators to be specialized. Nevertheless, they should learn how to generate realistic images because the STGS isolates a generator from others by a categorical sampling. The gating networks learn the type of $\mathbf{z}$ that best suits a certain generator and keep assigning similar ones to the generator. The generators learn that specializing on a particular subset of data distribution helps to obfuscate the discriminator by generating more realistic images. As the training iterations proceed, the generators converge to different local clusters of data distribution. 

\paragraph{Effect of Load-Balancing on Specialization} Another important aspect in training MEGAN is determining the hyperparameter $\lambda$ for the load-balancing regularization. A desired outcome from the assignment module is a logit vector \textit{\textbf{l}} with high variance among its elements, while maintaining the training of generators in a balanced manner. Although the load-balancing regularization is designed to balance workloads between generators, it slightly nudges the assignment module to yield a logit vector closer to a uniform distribution. Thus we observe when an extremely large value is set for $\lambda$ (e.g., 1000), the logit values follow a uniform distribution. It is not a desired consequence, because a uniform distribution of $l$ means the gating networks failed to properly perform the generator assignment, and the specialization effect of generators is minimized. To prevent this, we suggest two solutions.

The first solution is to obtain an optimal value of $\lambda$ where training is stable, and the logit values are not too uniform. It is a simple but reliable remedy, for finding the optimal $\lambda$ is not demanding. The second possible solution is to increase $\lambda$ when the logit values follow a uniform distribution. Most of our experiments were performed by the first method in which we fix $\lambda$, because a stable point could be found quickly, and it allows us to focus more on the general capability of the model.   
 
\paragraph{Data Efficiency} Some may claim that our model lacks data efficiency because each generator focuses on a small subset of a dataset. When trained with our algorithm, a single generator is exposed to a smaller number of images, because the generators specialize in a certain subset of the images. However, it also means that each generator can focus on learning fewer modes. Consequently, we observed that our model produces images with an improved quality, as described in detail in Section~\ref{sec: experimental results}.      

\section{Experiment Details}
In this section, we describe our experiment environments and objectives. All the program codes are available in https://github.com/heykeetae/MEGAN.

\subsection{Experiment Environments} \label{sec:ex_en}
We describe the detailed experiment environments in this section, such as baseline methods, datasets, etc.

\paragraph{Underlying GANs}
We apply our algorithm on both DCGAN and WGAN-GP (DCGAN layer architecture) frameworks, chosen based on their stability and high performance. The experiments consist of visual inspections, visual expertise analysis, quantitative evaluations, and user studies for generalized qualitative analyses.

 \begin{figure}[t]
\centering
\includegraphics[width=0.49\textwidth]{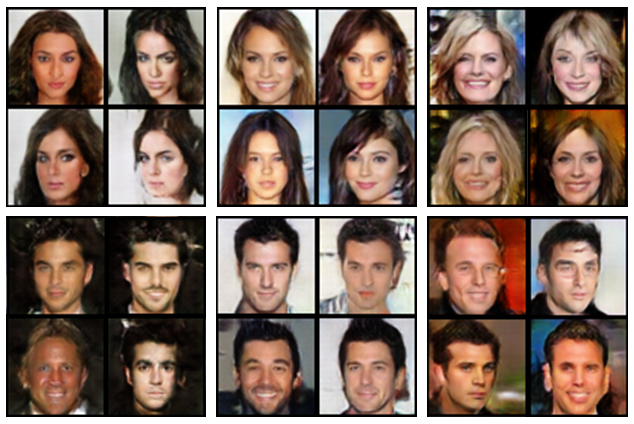}
\caption{\textbf{Visual Inspection}; CelebA dataset, 64x64 samples from MEGAN with each block of four images generated by the same generator. Noticeable differences between each block indicate that different generators produce images with different features.}\label{fig:celeb}
\end{figure}

\paragraph{Baseline algorithms} We compared our quantitative results with many state-of-the-art GAN models such as BEGAN, LSGAN, WGAN-GP, improved GAN (-L+HA)~\cite{salimans2016improved}, MGAN and SplittingGAN. AdaGAN could not be included for our evaluation, because the official code for AdaGAN does not provide a stable training for the datasets we used.  

\paragraph{Datasets} We used three datasets for our evaluation: CIFAR-10, CelebA, and LSUN. CIFAR-10 has 60,000 images from 10 different object classes. CelebA has 202,599 facial images of 10,177 celebrities. LSUN has various scenic images but we evaluated with the church outdoor subset, which consists of 126,227 images.

\paragraph{Evaluation Metric} We evaluated our model on two standard metrics to quantitatively measure the quality of the generated images: inception score (IS) and multiscale structural similarity (MS-SSIM). The IS is calculated through the inception object detection network and returns high scores when \textit{various} and \textit{high-quality} images are generated. MS-SSIM is also a widely used measure to check the similarity of two different images. We generated 2,000 images and checked their average pairwise MS-SSIM scores. The lower the score is, the better is the algorithm in terms of diversity.

\paragraph{User Study} We also conducted web-based user studies. In our test website,\footnote{\url{http://gantest.herokuapp.com} - a test run can be made by entering the following key: 5B3309} randomly selected real and fake images are displayed, and users are asked to downvote images that they think are fake. Nine images were provided per test and users iterate the test for 100 times. Regarding the CelebA dataset, we observed that the participants were good at detecting the generated facial images of the same race. Therefore, we diversified the ethnicity in our user groups by having thirty participants from three different continents. 

\paragraph{Hyperparameters} We tested the following hyperparameter setups: the number of generators $n$ as 3, 5, and 10; the mini-batch size $b$ = $64$; annealing temperature $\tau = 0.5\exp^{(-0.001\times{iter})}$~\cite{jang2016categorical} where $iter$ denotes the iteration number as in the Algorithm~\ref{alg:gan} ; load-balancing parameter $\lambda$=100; and feature vector $\mathbf{f}_i$ of dimension $k = 8192$ for CIFAR-10 and $16384$ for both LSUN and CelebA.

\section{Experimental Results}
\label{sec: experimental results}

\begin{figure}[t]
\centering
\includegraphics[width=0.49\textwidth]{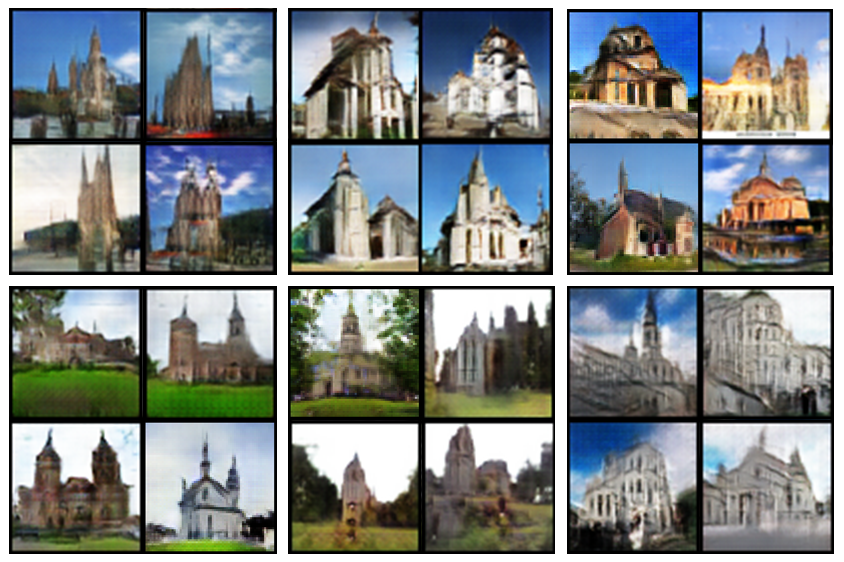}
\caption{\textbf{Visual Inspection}; LSUN-Church outdoor dataset, 64x64 samples from MEGAN with each block of four images generated by the same generator. Distinguishable features include the church architectural style, the location, and the cloud cover.}\label{fig:lsun}
\end{figure}

\subsection{Evaluation on Specialization}
We describe our results based on various visual inspections.

\begin{figure}[t]
\centering
\includegraphics[width=0.5\textwidth]{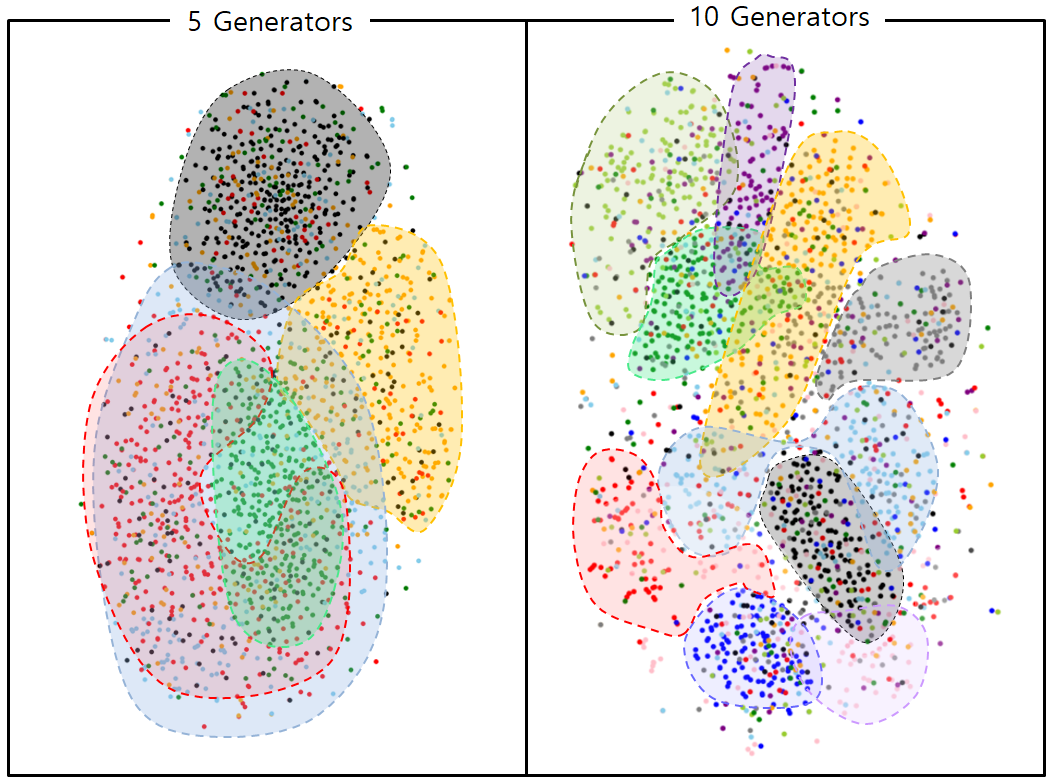}
\caption{\textbf{Visual Expertise Analysis}; 2,000 images are generated by MEGAN on CIFAR-10 dataset and feature vectors of those images are extracted from the \textit{relu4\_2} layer of VGG-19 networks. All 2,000 feature vectors are visualized in a two-dimensional space by the t-SNE algorithm.}\label{fig:visual}
\end{figure}

\paragraph{Visual Inspection}
Throughout our evaluations, each generator is found to learn different context and features, for at least up to 10 generators that we have inspected. The decision to assign a particular subset of data to a particular generator is typically based on visually recognizable features, such as background colors or the shape of the primary objects. Figs.~\ref{fig:celeb} and~\ref{fig:lsun} show the generated samples drawn from different generators trained with 10 generators on CelebA and LSUN-Church outdoor, respectively. Each of the block of four images are from the same generator. We chose six generators that have learned the most conspicuous and distinctive features readily captured even by the human eyes. 

 All four images share some features in common, while having at least one distinguishing characteristic from other blocks of images. For instance, the top-left celebrities in Fig.~\ref{fig:celeb} have black hair without noticeable facial expressions. On the contrary, the top-right celebrities have light-colored hair with smiling faces. Among the samples from LSUN in Fig.~\ref{fig:lsun}, we also detected distinguishing patterns specific to each generator.

\paragraph{Visual Expertise Analysis}
If the model learns properly, a desirable outcome is that each generator produces images of different features. We generated 2,000 CIFAR-10 images from MEGAN trained for 20 epochs, fed them to a pretrained VGG19 network, and extracted the feature vectors from the \textit{relu4\_2} layer. Subsequently, the 8,192-dimensional feature vectors are reduced to two-dimensional vectors using the t-SNE algorithm~\cite{maaten2008visualizing}. Fig.~\ref{fig:visual} shows the results. Each two-dimensional vector is represented as a dot in the figure, and samples from the same generator are of the same color. The colored shades indicate the clusters of images that are generated by the same generator. We tested MEGAN with 5 generators and 10 generators, confirming that each generator occupies its own region in the feature vector space. It is noteworthy that they overlap in the figure owing to dimensionality reduction for visualization purposes. In the original 8192-dimension space, they may overlap much less.

\begin{table}[t]
\begin{center}
\caption{\textbf{\label{table:cifar}Inception Score on CIFAR-10} (trained without labels)}
\begin{tabular}{lcr}  
\toprule
\textbf{Method}    & \textbf{Score}  \\
\midrule
DCGAN      & $6.16 \pm 0.06$       \\
Improved GAN (-L+HA)& $6.86 \pm 0.07$  \\
WGAN-GP (Resnet) & $7.86 \pm 0.07$ \\
SplittingGAN & $7.90 \pm 0.09$ \\
AdaGAN & Not being properly trained\\
\textbf{MGAN} & $\textbf{8.33}\  \mathbf{\pm}\ \textbf{0.10}$\\
\textbf{MEGAN (DCGAN)} & $\textbf{8.33}\ \mathbf{\pm}\  \textbf{0.09}$ \\

\bottomrule
\end{tabular}
\end{center}
\end{table}

\begin{table}[t]
\begin{center}
\caption{\label{table:celeb}\textbf{MS-SSIM Score on CelebA}}
\begin{tabular}{lccc}  
\toprule
\textbf{Method}    & $n$ &\textbf{CelebA} & \textbf{LSUN} \\ 
\midrule
BEGAN  &1& $0.4636 \pm 0.019$  & $0.1969 \pm 0.024$  \\
DRAGAN &1& $0.3711 \pm 0.020$ &  $0.1733 \pm 0.015$  \\
LSGAN  &1& $0.3487 \pm 0.019$ &  $0.1067 \pm 0.016$  \\
\\[-0.9em]
MGAN &5& $0.2611 \pm 0.021$ & $0.1142 \pm 0.015$\\
MGAN &10& $0.2816 \pm 0.012$ & $0.1055 \pm 0.013$\\
\\[-0.9em]
MEGAN &3& $0.2818 \pm 0.020$ & $0.1024 \pm 0.015$\\
MEGAN &5& $\mathbf{0.2470 \pm 0.024}$ & $\mathbf{0.0997 \pm 0.021}$\\
MEGAN &10& $0.2665 \pm 0.027$ & $0.1085 \pm 0.014$\\
\bottomrule
\end{tabular}
\end{center}

\caption{\textbf{\label{table:user}: User study results}}
\scalebox{0.85}{
\begin{tabular}{p{1.98cm}rclp{0.01cm}rcl}
\toprule
       & & \textbf{CelebA} & & & & \textbf{LSUN} & \\
\midrule
\\[-0.9em]
 & Avg     & Min     & Max   &  & Avg  & Min & Max \\
 \\[-0.9em]
 \hline
 \\[-0.8em]
BEGAN & \textbf{0.70} & 0.50 &  0.89  & & 0.73   &  0.49 &0.93 \\
DRAGAN& 0.91 &   0.71 &0.98   & & 0.81  &  0.50  &  0.96    \\
LSGAN & 0.88 &    0.71 & 0.96  &   & 0.59  &   0.36  &  0.91   \\
WGAN-GP& 0.82 &    0.70 &  0.96 & &0.58 &  0.33  &  0.85   \\
\begin{tabular}[c]{@{}c@{}}MEGAN\\ ($n=3$)\end{tabular} & 0.76 &   0.58 &   0.95 & & \textbf{0.49}     &   0.20   &  0.71   \\
\begin{tabular}[c]{@{}c@{}}MEGAN\\ ($n=5$)\end{tabular} & 0.73        &    0.57 &     0.93 & & 0.61     &  0.40   &  0.81 \\
\begin{tabular}[c]{@{}c@{}}MEGAN\\ ($n=10$)\end{tabular} & 0.74 &     0.60  & 0.92 & & 0.58  &  0.28   &   0.92   \\
\bottomrule
\end{tabular}}
\end{table}

\subsection{Quantitative analysis}
We introduce our quantitative experiment results using the inception score, MS-SSIM, and user study.

\paragraph{CIFAR-10}
Table~\ref{table:cifar} lists the inception scores (Section~\ref{sec:ex_en}) of various models on CIFAR-10. MEGAN trained on the DCGAN records an inception score of 8.33 --- our MEGAN shows a slightly better variance, i.e., 0.09 in MEGAN vs.0.1 in the MGAN. The official code for the AdaGAN does not provide a stable training for the CIFAR-10 dataset, and its inception score is not comparable to other baseline methods.



\paragraph{CelebA and LSUN-Church outdoor}
The MS-SSIM scores (Section~\ref{sec:ex_en}) measured for CelebA and LSUN-Church outdoor are reported in Table~\ref{table:celeb}. As the MS-SSIM scores of the baseline models are missing in their papers, we evaluate them after generating many samples using their official codes. In this experiment, MEGAN is trained to minimize the WGAN-GP loss function that was found to perform best in our preliminary experiments. We report that MEGAN outperforms all baseline models in terms of the diversity of generated images, as shown by its lowest MS-SSIM scores. Notably, MEGAN with five generators achieve the lowest MS-SSIM scores for both datasets.

\paragraph{User Study}
Table~\ref{table:user} shows the result of the web-based user study on CelebA and LSUN-Church outdoor datasets. The score is computed by dividing the number of downvoted fake images by the total number of fake images shown to users. Thus, a low score indicates that users struggle to distinguish generated images from real images. For both datasets, MEGAN records competitive performance, and especially for LSUN-Church outdoor it outperforms all the baseline models.   

In conjunction with the previous MS-SSIM results, MEGAN's low detection rates indicate that it can generate more diverse images in better quality than other baseline methods. We observed that BEGAN achieves the lowest detection rate for the CelebA dataset, in exchange for the low diversity of generated images, as indicated by the high MS-SSIM score of BEGAN in Table~\ref{table:celeb}.  

\section{Conclusion}
This paper proposed a novel generative adversarial networks model called MEGAN, for learning the complex underlying modalities of datasets. Both our quantitative and qualitative analyses suggest that our method is suitable for various datasets. Future work involves extending our algorithm to other variants of GANs and a broader range of generative models.


\bibliographystyle{named}
\bibliography{ijcai18}

\end{document}